# A Machine Learning Model for Predicting, Diagnosing, and Mitigating Health Disparities in Hospital Readmission

Shaina Raza, PhD

Shaina.raza@utoronto.ca

## Abstract

The management of hyperglycemia in hospitalized patients has a significant impact on both morbidity and mortality. Therefore, it is important to predict the need for diabetic patients to be hospitalized. However, using standard machine learning approaches to make these predictions may result in health disparities caused by biases in the data related to social determinants (such as race, age, and gender). These biases must be removed early in the data collection process, before they enter the system and are reinforced by model predictions, resulting in biases in the model's decisions. In this paper, we propose a machine learning pipeline capable of making predictions as well as detecting and mitigating biases in the data and model predictions. This pipeline analyses the clinical data and determines whether biases exist in the data, if so, it removes those biases before making predictions. We evaluate the performance of the proposed method on a clinical dataset using accuracy and fairness measures. The findings of the results show that when we mitigate biases early during the data ingestion, we get fairer predictions.

## 1 Introduction

Hyperglycemia is the medical term for having high blood glucose levels (blood sugar) [1]. Hemoglobin A1c (HbA1c) is a blood test that is used to help diagnose and monitor people with diabetes [2]. When diabetic patients come to the emergency department, their HbA1c level should be determined. The outcome can then be used to assess the patient's current glycemic control and the urgency of needed follow-up (e.g., hospitalization).

The management of hyperglycemia in hospitalized patients has a significant impact on morbidity and mortality [3]. Many hospitals have begun to adhere to formalized guidelines and set tight glucose targets for critical patients in their intensive care units (ICUs). However, the same protocols are not followed for most admissions due to health disparities in the system.

A health disparity [4] is a difference or gap in health status or health care supply and access that is frequently associated with social, economic, and environmental adversity. Health inequalities impact groups of people who experience systematic barriers to receiving health care services due to health disparities [5].

According to research [1], examining historical patterns of diabetes care in hospitalized diabetic patients may lead to improvements in patient safety. Some studies [6] show that measuring (HbA1c) is also associated with lower readmission rates in hospitalized patients. We will continue this line of research (evaluating the effect of HbA1c testing on hospital readmission rates) but we additional evaluate if there are racial/ethnic disparities that may affect the hospitalization rate. The goal is of this research to reduce the disparities in readmission rates of diabetic patients based on gender/race/ethnicity.

The potential impact of big data in health science cannot be ignored. It is also quite evident now that Machine Learning (ML) algorithms developed from big data could worsen or perhaps create new forms of disparities [7]. Bias in ML literature is usually defined in terms of the dataset or model—bias in labelling, sample selection, data retrieval, scaling and imputation, or model selection [8]. These biases can be mitigated at three stages: early, mid, and late [8] of a software development life cycle. The early stage (pre-processing) would be to reduce bias by manipulating the training data before training the algorithm [9]. The mid-stage (in-processing) would be to de-bias the model itself [10], which is also framed as an optimization problem. The late stage (post-processing) refers to reducing the biases by manipulating the output predictions after training [11]. Prior research [12, 13] has shown that missing the chance to detect biases at an early stage can make it difficult for an ML pipeline to mitigate them later on.

An ML pipeline is composed of several steps that facilitate the automation of ML workflows [14]. Usually, these pipelines are designed to input some data as features and generate a score that predicts, classifies, or recommends future outcomes. In this work, we propose a fair ML pipeline, designed exclusively for mitigating data biases and leveraging fairness in the healthcare setting. In this paper, we propose a fair ML pipeline that can take a large clinical data and de-biases it early on. We summarize our contributions as:

1. We create a fair ML pipeline that can predict readmission rates for individuals with diabetes mellitus while taking care to minimize health disparities among vulnerable population groups during predictions.
2. We build the fair ML pipeline according to the widely accepted data science pipeline structure [15], intending to make it easier for practitioners to use. This pipeline is intended to reduce health disparities; however, given another dataset, this design can be used to reduce biases in that domain; the only requirement is that the pipeline be trained on new data;
3. We conduct on a benchmark dataset to determine how predictions are impacted before and after de-biasing the data. our findings show that when we mitigate biases to introduce fairness, we also lose some level of accuracy.

The rest of the paper is organized as:

Section 2 is the literature review, section 3 is the proposed methodology, section 4 is the experimental setup, section 5 is the discussion and section 6 is the conclusion.

## 2 Literature Review

Bias is defined as an inclination or prejudice for or against one person or group, especially in an unfair manner [16]. In most use cases, it is the underlying data that introduces biases in the ML models. There are numerous examples of biases in ML models due to data, such as Amazon's hiring algorithm[1] that favoured men, Facebook targeted housing advertising that discriminated based on race and colour[2] and a healthcare algorithm [17] that exhibited significant racial biases in its recommendations.

Fairness [8] is a multi-faceted concept that varies by culture and context. There exist at least 21 mathematical definitions for fairness in politics [18]. A decision tree[3] on different definitions of fairness is provided by the University of Chicago. Most of the definitions of fairness focus on either individual fairness (treating similar individuals fairly) or group fairness (equitably distributing the model's predictions or outcomes across groups) [8, 19]. Individual fairness aims to ensure that statistical measures of outcomes are equal for statistically similar people. Group fairness divides a population into distinct groups based on protected characteristics, to ensure that statistical measures of outcomes are comparable across groups.

***Fairness algorithms***: In the research of AI and ML fairness [8, 19, 20], the bias mitigation algorithms are categorized into three broad types: (1) pre-processing algorithms; (2) in-processing algorithms; and (3) post-processing algorithms. These algorithms are briefly discussed below.

*Pre-processing algorithms*: The pre-processing algorithms attempt to learn a new representation of data by removing the information associated with the sensitive attribute while retaining as much of the actual data as possible. This technique manipulates the training data before training the algorithm. Well-known pre-processing algorithms are:

- A reweighting [9] algorithm that generates weights for the training samples in each (group, label), without changing the actual feature or label values.
- The learning fair representations [21] algorithm discovers a latent representation by encoding the data while concealing information about protected attributes. Protected attributes are those attributes that divide a population into groups whose outcomes should be comparable (such as race, gender, caste, and religion) [8].
- The disparate impact remover [22] algorithm modifies feature values to improve group fairness while keeping rank ordering within groups.
- Optimized pre-processing [23] algorithm learns a probabilistic transformation that edits data features and labels for individual and group fairness.

Usually, pre-processing techniques are easy to use as the modified data can be used for any downstream tasks without requiring any changes to the model itself.

---

[1] us-amazon-com-jobs-automation.
[2] facebook-ads-housing-discrimination-charges
[3] http://aequitas.dssg.io/static/images/metrictree.png

*In-processing algorithms*: In-processing algorithms penalize the undesired biases from the model, to incorporate fairness into the model. The in-processing technique influences the loss function during the model training to mitigate biases. In the past, the in-processing algorithms [24, 25] have been used to provide equal access to racially and ethnically diverse group. Some of the example in-processing algorithms are listed below:

- Prejudice remover [24] augments the learning objective with a discrimination-aware regularization term.
- Adversarial De-biasing [26] algorithm learns a classifier to maximize the prediction accuracy while decreasing an adversary's ability to deduce the protected attribute from the predictions.
- Exponentiated gradient reduction [27] algorithm breaks down fair classification into a series of cost-sensitive classification problems, returning a randomized classifier with the lowest empirical error subject to fair classification constraints.
- Meta fair classifier [25] algorithm inputs the fairness metric and returns a classifier that is optimized for the metric.

The in-processing technique is model or task-specific, and it requires changes within the algorithm, which is not always a feasible option.

*Post-processing algorithms:* The post-processing algorithms manipulate output predictions after training to reduce bias. These algorithms can be applied without retraining the existing classifiers (as in in-processing). Some of the post-processing algorithms are:

- Reject option classification [28] algorithm gives favourable outcomes (labels that provide an advantage to an individual or group, e.g., being hired for a job, not fired) to unprivileged groups and unfavourable outcomes (not hired for a job, fired) to privileged groups (has historically been at a systemic advantage).
- Equalized odds [29] algorithm changes the output labels to optimize equalized odds through linear programming.
- Calibrated equalized odds [11] optimize score outputs to find probabilities with which to change output labels with an equalized odds objective.

Usually, the post-processing technique requires access to protected attributes late in the pipeline.

Recently, the software engineering community has also started to work on fairness in ML, specifically fairness testing [30]. Some work has been done to develop automated tools, such as AI Fairness 360 [20], FairML [31], Aequitas [32], Themis-ML [33], and FairTest [34], that follows a software development lifecycle. These fairness toolkits are discussed as:

- FairML [31] is a toolkit that uses few ranking algorithms to quantify the relative effects of various inputs on a model's predictions, which can be used to assess the fairness in models.

- FairTest [34] is a python package that learns a special decision tree that divides a user population into smaller subgroups with the greatest possible association between protected (such as gender, race, and other characteristics deemed sensitive features) and algorithm outputs.
- Themis-ml [33] is a python library that implements several state-of-the-art fairness-aware methods that comply with the sklearn API.
- AIF360 [20] consists of several fairness metrics for datasets and state-of-the-art ML models to mitigate biases in the datasets and models.

***Fairness metrics****:* To ensure fairness, the ML community has also proposed various fairness metrics. For example, the disparate impact ratio compares the rate at which an underprivileged group (groups having systematic biases e.g., females) receives a particular outcome compared to a privileged group (groups with systematic advantages, e.g., males) [22]. In some works [8], the number of positives and negatives for each individual or group, a difference of means, and an odds ratio are also computed to measure fairness.

This work focus on a pipeline structure to detect and mitigate biases in hospitalization data. This pipeline is developed with re-suability and generalizability in mind.

## 3 Proposed Methodology

In this section, we present the proposed methodology.

### 3.1 Key terms

Some of the preliminaries and key terms used in this work are given.

- *Bias* is a type of systemic error.
- A *protected attribute* divides a population into groups, for example, race, gender, caste, and religion.
- A *privileged* value of a protected attribute denotes a group that historically has a systematic advantage, for example, male gender, white race.
- An *underprivileged* group faces prejudice based on systematic biases such as gender, race, age, disability, and so on.
- *Group fairness* means that the protected attribute receives similar treatments or outcomes.
- *Individual fairness* means that similar individuals receive similar treatments or outcomes.
- *Equalized odds* [35] is a statistical notion of fairness that ensures classification algorithms do not discriminate against protected groups.
- *Fairness* or *fair* [16] is a broad term used in ML and generally refers to the process of undoing the effects of various biases in the data and algorithms.

Some of the notations used in this work are in Table 1:

**Table 1**: Notations used

| Symbol | Description |
|---|---|
| $y \in 0{,}1$ | Actual value or outcome |
| $\hat{y} \in 0{,}1$ | Predicted value or outcome |
| $Pr(\hat{y}_i = 1)$ | Probability of $(\hat{y}_i = 1)$ for observation $i$ |
| $g_i, g_j$ | Identifier for groups based on protected attribute. A group is either privileged or unprivileged. |

### 3.2 Pipeline structure

Our fair ML pipeline is shown in Figure 1.

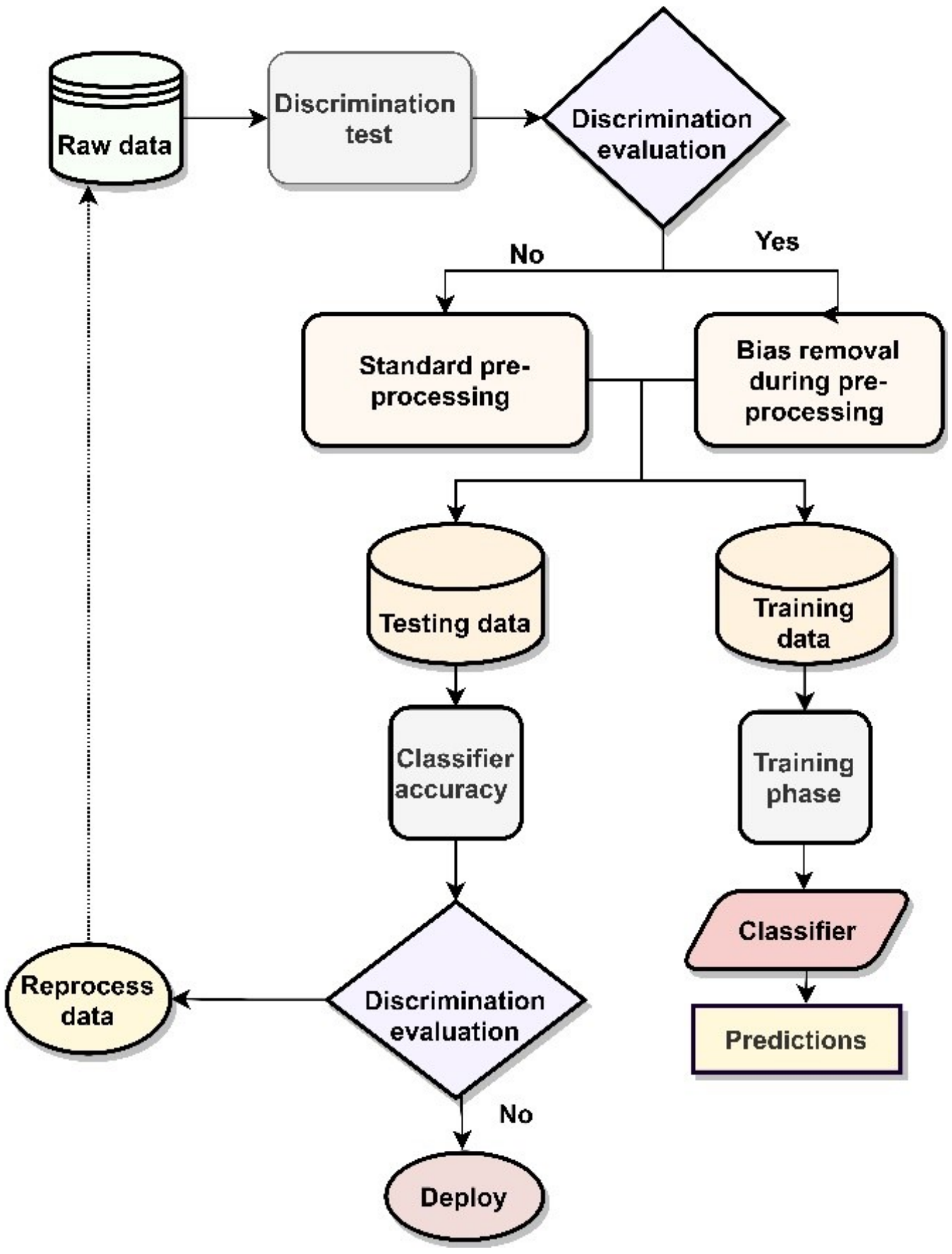

**Figure 1:** Fair ML pipeline for hospital readmissions

We develop a fair ML pipeline to examine, report and mitigate discrimination and biases in the hospital readmission of diabetic patients based on HbA1c measurement [6]. We use the same classification strategy as in the original paper [6] to determine the likelihood of readmission (target variable) based on the patient's records (patient is readmitted within 30 days of discharge, HbA1c measurements, HbA1c > 8%, medication) to determine the outcome. However, our work contributes to determining whether hospitalization readmissions are equitable for all groups based on gender, age and race.

As shown in Figure 1, we start with the raw data, which in this case is diabetic patient hospitalization data [6]. The data is first given to the discrimination test unit, which determines whether certain privileged groups (e.g., white, male, adult) are given a systematic advantage and certain unprivileged groups are given a systematic disadvantage. This applies to characteristics such as race, gender and age. If the discrimination test shows the existence of biases which we determine through fairness metrics (defined in Section 4.3), we use a pre-processing fairness algorithm to reduce or remove those biases. We manipulate the training data before training the algorithm to remove biases. If there are no biases, we use the standard ML pre-processing technique. After pre-processing (either bias removal or standard preprocessing), the data is separated into training and testing data. Then, we perform the classification task and make predictions. The test data is used to determine the accuracy of the classification. We repeat the discrimination test on the test data, which we assume is bias-free at this point. If the discrimination evaluation reveals no biases, we put the model into production for use. Otherwise, we reprocess the data to correct any biases that were missed during the bias pre-processing stage and send it back to the pipeline.

### 3.3 Binary classification and bias removal

Formally, we define the classification and bias removal problem as:

Given a dataset $D = (p, x, y)$ consisting of protected attributes $p$ (race, gender, age), remaining attributes $x$ (features other than $p$), and binary class $y$ (patient is 'readmitted' or 'not readmitted'), the goal is to have a transformed dataset $D' = (p, x', y)$ ($x'$ is a modified set of attributes) that is guaranteed to be having no health disparities. The goal is to change only the remaining attributes $x$, leaving $y$ unchanged from the original data set to preserve predictions.

***Classification algorithm***: We use the Logistic Regression (LG) [36] as the classifier. LG estimates the probability of an event occurring, such as readmitted or not, based on the independent variables in the dataset. Since the outcome is a probability, the dependent variable is bounded between 0 and 1. We chose LG over other robust models including more recent deep neural networks, due to the numeric and categorical nature of our data.

***Fairness algorithm***: In this work, we use Reweighting [9] as a preprocessing technique to introduce fairness. Reweighting method [9] weighs the examples in each (group, label) combination differently to ensure fairness before classification. Here by groups, we mean grouping based on protected attributes (gender, race, age). This technique allows us to intervene in the data modelling process. We assign different weights to the examples (in data) based on their categories of the protected attributes (based on gender, race, and age), which removes the biases from the training dataset. The weights are based on frequency counts. The outline of this algorithm is as:

- Sample the training set uniformly to ensure a pre-specified number of samples per class.

- Look at the protected attribute and on the real label.
- Calculates the probability of assigning a favorable label (*y*=1) assuming the protected attribute and *y* are independent. If there is bias, they will be statistically dependent.
- Divide the calculated theoretical probability by true to create the weight.
- Create a weight vector for each observation in data using 2 vectors (protected variable and y).
- Weigh the examples in each (group, label) combination differently to ensure fairness before classification.

We measure the health disparity between different groups in terms of a Disparity Impact (DI) ratio [37]. We use the DI fairness ratio to define fairness in this (binary) classification setting. An ML model has DI when its performance changes across groups defined by protected attributes [38]. We define DI in equation 1 as:

$$\frac{Pr(\hat{y} = 1|\ g_1)}{Pr(\hat{y} = 1|\ g_2)} \quad (1)$$

This calculation measures the percentage of unprivileged groups ($g_1$) who received a favourable outcome is divided by the percentage of privileged groups ($g_2$) who received a favourable outcome. DI has its origin in legal fairness considerations [22] that often use the four-fifths rule [39], which means if the unprivileged group receives a positive outcome of less than 80% of their proportion of the privileged group, this is a DI violation

The purpose of all this methodology is to guarantee that, given *D*, any classification algorithm aiming to predict $\hat{y}$, will not have DI. We are interested in the outcomes to ensure that protected attributes do not suffer from biases.

# 4 Experimental Setup

## 4.1 Dataset

We use the "Diabetes 130-US hospitals Data Set for the years 1999-2008" [40] in this research. The data contains attributes, such as patient number, gender, age, race, admission type, time in hospital, the medical specialty of admitting physician, HbA1c test result, diagnosis medications and so. All the electronic medical records are de-identified by the dataset providers [6] and consist of around 50 features that we also use. The dataset features used in this work are in Table 1:

**Table 2**: Dataset features used in this work

| Feature | Type | Description and values |
|---|---|---|
| Patient number | Numeric | Unique identifier of a patient |
| Race | Nominal | Caucasian, Asian, African American, Hispanic, and other |
| Gender | Nominal | Male, female |
| Age | Nominal | Grouped in 10-year intervals: [0, 10), [10, 20), ..., [90, 100) |
| Weight | Numeric | Weight in pounds. |
| Admission type | Nominal | Emergency, urgent, elective, newborn, and not available |
| Discharge disposition | Nominal | Discharged to home, expired, and not available |
| Admission source | Nominal | Physician referral, emergency room, and transfer from a hospital |
| Time in hospital | Numeric | Days between admission & discharge |
| Number of lab procedures | Numeric | Number of lab tests performed during the encounter |
| Glucose serum test result | Nominal | Indicates the range of the result or if the test was not taken. Values: “>200,” “>300,” “normal,” and “none” if not measured |
| A1c test result | Nominal | was greater than 8%, “>7” if the result was greater than 7% but less than 8%, “normal” if the result was less than 7%, and “none” if not measured. |
| Diabetes medications | Nominal | Indicates if there was any diabetic medication prescribed. Values: “yes” and “no” |
| Readmitted | Nominal | 30 days, “>30” if the patient was readmitted in more than 30 days, and “No” for no record of readmission |

### 4.2 Baseline methods

Our proposed ML pipeline has two important tasks: 1) to classify (i.e., to predict the hospitalization re-admission for diabetes patients), and 2) to determine the existence of the biases in the prediction. So, we use the following baseline methods based on the task to perform.

To evaluate the classification task, we use the Logistic Regression (LG) as the baseline method.

To evaluate the fairness, we use the Disparate Impact Remover (DIR) [22] as the pre-processing fairness method, Adversarial Debiasing (AD) [26] as the in-processing fairness method, and Calibrated Equalized Odds (CEO) [21] as the post-processing method. The selection of these baselines is based on the usage of these methods in related works [8, 41]. The details of pre, in and post-processing methods are given in Section 2 (Literature Review).

### 4.3 Evaluation protocol, metrics, and strategy

**Evaluation protocol**: First, we define the protected attributes: {‘age’, ‘gender’, ‘race’}. Protected attributes are those attributes that divide a population into groups whose outcomes should be comparable (such as race, gender, caste, and religion) [8]. Based on the protected attributes, we define the privileged and unprivileged groups. A *privileged* value of a protected attribute denotes a group that historically has a systematic advantage, for example, male gender, white race [22]. An *underprivileged* group faces prejudice based on systematic biases such as gender, race, age, disability, and so on [22]. In this work, we define the privileged and unprivileged groups as:

- *Privileged groups*: {‘age’: 1 (older than or equal to 25 AND less than 65); ‘gender’: 1 (Male); ‘race’: 1 (‘non-Black’)}

- *Unprivileged groups*: {'age': 0 (younger or equal to 25 AND above 65 ); 'gender': 0 (Female); 'race': 0 ('Black')}.

We define these groups using information from the United Nations Development Programme (UNDP) regarding marginalized groups [42]. Also, a preliminary analysis of the dataset used in this study revealed that hospitalizations for the protected groups are distinct. Consequently, we define these groups and then assess the distribution of hospitalizations in this study via our experiments.

**Evaluation metrics**: We use the following evaluation metrics following the standard protocols in related works [8, 20].

1) *Disparate Impact (DI)* [34] is an evaluation metric to evaluate fairness.

DI compares the proportion of individuals that receive a positive output for two groups: an unprivileged group and a privileged group. The industry standard for DI is a four-fifths rule [39], which means if the unprivileged group receives a positive outcome of less than 80% of their proportion of the privileged group, this is a disparate impact violation. An acceptable threshold should be between 0.8 and 1.25, with 0.8 favouring the privileged group, and 1.25 favouring the unprivileged group [39]. Mathematically, it can be defined in equation 2:

$$DI = \frac{\frac{num_positives(privileged=False)}{num_instances(privileged=False)}}{\frac{num_positives(privileged=True)}{num_instances(privileged=True)}} \tag{2}$$

where *num_positives* are the number of individuals in the group: either privileged=False (unprivileged), or privileged=True (privileged), who received a positive outcome. The *num_instances* are the total number of individuals in the group.

2) *Average odds difference (avg_odd)* [20] is a fairness metric, it is the average of the difference in false positive rates and true positive rates between unprivileged and privileged groups. A value of 0 implies both groups have equal benefits.
3) *Equal opportunity difference (eq_opp)* [20] is a fairness metric, it is the difference in true positive rates between unprivileged and privileged groups. A value of 0 implies both groups have equal benefits.
4) *Balanced accuracy (acc)* [42] is an accuracy metric, it evaluates the accuracy of a classifier, and a value close to 1 is a good value. It is particularly useful when the classes are imbalanced, i.e., when one class appears significantly more frequently than the other.

The *DI, avg_odd* and *eq_opp* are fairness metrics while the *acc* is the accuracy metric.

The intuition behind using the fairness metrics is to correct the biases in automated decision processes based on ML models. For example, to evaluate the model for the biases related to gender, race, disability and more. We use these metrics to check whether a classifier produces the same result for one individual/group as it does for another individual/group who is identical to the first.

**Evaluation Strategy**: These experiments are conducted in a two-fold manner: (1) evaluation before de-biasing, and (2) evaluation after de-biasing. The outline of evaluation is given below:

1. We split the dataset into training and test sets. We train the pipeline on the train set and test on the test set.
2. We use LG as the classifier to predict the outcomes (hospitalization re-admission) and evaluate the classifier performance task (this is classification evaluation).
3. We quantify the existence of the biases using fairness evaluation metrics on model predictions, based on the information encoded among different groups (privileged/ unprivileged) (this is a fairness assessment task).
4. We de-bias the data using each of the baseline's specific methodology (pre-processing, in-processing and post-processing) (this is a bias mitigation task). The result is a transformed dataset.
5. We again quantity the existence of the biases on the transformed dataset using fairness metrics (this is fairness assessment).
6. Finally, we evaluate the accuracy of the classifier on the transformed data (this is a classification assessment).

The overall experimental setup is given in Table 3.

**Table 3**: Experimental setup

| | |
|---|---|
| **Dataset** | Diabetes 130-US hospitals for years 1999-2008 |
| **Protected attributes** | Race, Gender, Age |
| **Privileged class** | White, Male, Adult (25-65 years) |
| **Unprivileged class** | Non-white, Female, not Adult |
| **Classifier** | Logistic Regression (LG) |
| **Bias mitigation method** | Disparate Impact Remover (DIR), Adversarial Debiasing (AD) [26]and Calibrated Equalized Odds (CEO) |

### 4.4 Results

We present the results on different baseline methods in Table 4.

**Table 4**: Classification and Fairness results

| | ***Before de-biasing (original data)*** | | | |
|---|---|---|---|---|
| **Method** | **acc** | **avg_odd** | **eq_opp** | **DI** |
| Classifier (LG) | **84.30%** | -0.2018 | -0.1182 | 0.750 |
| | ***After de-biasing (transformed data)*** | | | |
| **Method** | **Acc of classifier** | **avg_odd** | **eq_opp** | **DI** |
| DIR | 67.70% | -0.021 | -0.017 | 0.913 |
| AD | 65.53% | -0.042 | -0.022 | 0.835 |
| CEO | 62.88% | -0.044 | -0.025 | 0.132 |
| Ours | **80.15%** | -0.019 | 0.001 | 1.012 |

In Table 4, the evaluation before de-biasing shows the accuracy of the classifier (LG in this experiment) in classifying the outcomes. The fairness evaluation through fairness metrics (Average Odds, Equation Opportunity and Disparate Impact) is performed on the original dataset to see the existence of the biases in the data.

In the evaluation after-biasing, we use each of the method's fairness strategies to mitigate biases in different ways (DIR-pre-processing, AD-in-processing and COE-post-processing) to produce transformed data, and we again evaluate the fairness of the transformed data.

From the above results, we see that the classifier model with hyperparameter tuning gives us the best results (in terms of accuracy) on the original dataset. This shows how accurate the model predictions are compared to ground truth labels. However, our actual motive in this research is to increase the disparate impact fairness metric above 0.8 and to bring average odds and equal opportunity equal to 0.

We also observe in Table 4, that the classifier accuracy is decreased to about 4% in the transformed data, and the fairness is improved for each method. We see that our method shows the DI fairness of 1.012, which is quite above the 0.8 limit and better than the DI fairness of other methods. For other methods, the DI fairness also increased above 0.8, however, we see that the value is DI is still below 1.0 as in the case of AD and DIR, which means privileged groups are still favoured. For COE, we see the DI fairness is above 1.25, which means favouring the unprivileged groups.

We also observe in Table 4 that the Average Odds difference fairness for our approach has increased from their initial negative values to close to zero, which shows fairness is achieved. The fairness in terms of Average Odds for other models is also improved, but not as close to zero as returned by our method.

Our method also shows better fairness results in Equality of Opportunity. A value close to zero indicates that both groups have equality of benefit. Thus, we can say that we were successfully able to improve the fairness metrics after applying the Reweighing preprocessing Algorithm to mitigate the bias.

Overall, these results also indicate that when we achieve fairness, we lose some level of accuracy, which is understandable because, after debiasing, the sentence cannot be successfully detected as being originally biased, which aligns with the previous research [8, 20].

We also find that, in general, the pre-processing method (DIR) performs better than other methods. This is probably because when we remove the biases early in the model pipeline, as in pre-processing, we are feeding a transformed dataset that is quite free of biases and the impact of this bias mitigation strategy is reflected in the model predictions.

# 5 Discussions

## 5.1 Generalizability of proposed pipeline

The objective of this research is to develop a generalizable ML pipeline to make predictions for hospitalized patients and to classify and detect any biases before making the predictions. This pipeline is developed with generalizability and reusability in mind. By generalizability, it means that this pipeline can be used for any other healthcare or life sciences scenario [43], for example, to detect biases and make predictions for COVID-19 cases, for oncological purposes or so. And, reusability, means that the same pipeline can be used for other purposes with least or no code change. The only prerequisite for generalizability and reusability is to train the pipeline on the domain-specific data.

Since the focus of this study is hospitalized diabetic patients, we chose a dataset consisting of categorical data, based on the availability of a domain-specific option at the time. However, the same pipeline can be used to mitigate biases in other datasets consisting of various data types (such as texts, and numeric data types). However, to use this pipeline for detecting textual biases, we need to integrate a language model [44, 45] in the pipeline. The language models are trained on text corpora in one or many languages. These models are useful for a variety of problems in computational linguistics, such as speech recognition, natural language generation (generating more human-like text), part-of-speech tagging, parsing, information retrieval, and lately bias detection and mitigation tasks [46].

## 5.2 Machine learning models for classification and fairness

In this study, we predict outcomes using a straightforward classification model LG. Our objective is to examine the classification of the outcomes and their classification accuracy before and after de-biasing. There are numerous state-of-the-art classifiers, such as those from the Transformers, such as BERT [45], or others that can be found in AutoML[4]; however, our focus in this research is on mitigating biases in a simple dataset with numeric and categorical fields, so a simple classifier suffices the purpose. In the future, we also plan to test out alternative classifiers.

There are also some recent models [47, 48], such as Transformer-based models or other deep neural networks that are mostly used to de-bias textual content. Despite their computational excellence, we find that these methods are inadequate for mitigating bias in a simplistic dataset as we are using in this work. Our objective in this work is simplicity and debiasing, so the models, such as reweighting techniques, are adequate.

---

[4] https://www.ibm.com/cloud/watson-studio/autoai?

### 5.3 Bias mitigation during pre-processing

In this work, we use the pre-processing fairness technique to mitigate biases and introduce fairness in the results. Preprocessing techniques for bias mitigation is typically data-driven [49]. We choose to work with pre-processing technique because our preliminary analysis of the dataset revealed to us that certain characteristics of the training data may directly contribute to one protected or subgroup over the other. Due to this, we focus on modifying the training set by reweighting examples of data. By modifying the training data, we also find that the learned classifier's outputs is less biased. In future, we may try other fairness techniques, such as in-processing or post-processing, depending on the use case and the problem at hand.

### 5.4 Accuracy/Fairness Tradeoff:

When we introduce fairness in the results, we must recognize that this fairness does not come for free, in many cases, it may conflict with model accuracy [50]. For example, a model that is as accurate as possible (relative to the available ground truth) may discriminate against at least one subpopulation. Similarly, a model that is perfectly fair and equitable across all populations, such a model must be less accurate than one that does not take fairness into account. The behaviour of fair algorithms when deployed in the real world will not always match the results demonstrated in theory; therefore, understanding the relationship between fairness and accuracy is essential for having confidence in the models we select.

### 5.5 Limitations

A few limitations of this study and some future directions are given below:

1. There is no universally accepted definition of what constitutes bias and fairness [49]. As a future direction, we must first investigate different definitions of biases, especially, in the health domain to determine the fairness of data and algorithms.
2. We consider a small number of biases, we understand that there are many other types of health disparities, such as food, safety, health coverage and recently related to COVID-19, which may be overlooked in the research.
3. One of the research's limitations is the scarcity of labelled biased health-related data [51]. So far, we have used a benchmark dataset to assist us in detecting and mitigating bias. However, we would like to acquire more labelled data in the future to show how to mitigate biases on a broader dataset.

There is much work to do toward achieving fairness in the health domain, and we hope that others in the research community will contribute to this research.

## 6 Conclusion

One of the overarching goals of Healthy People 2030 [52] is to eradicate health disparities and develop health literacy among populations, which is also a motivation for this research. According to a review of studies [53, 54], ethnic minorities are disproportionately affected by the majority of diabetes complications. In this research, we try to find if the outcome (readmission to hospitalization) is favourable or unfavourable for different groups.

Through a pre-processing fairness approach, we mitigate biases and demonstrate fairness through extensive experiments. We find that prediction accuracy decreases during the de-biasing process. we compare our method with other fairness techniques (post-processing and in-processing) and find that pre-processing methods give us fairer predictions, which are validated through different fairness measures. We also get good accuracy when we de-bias the data before model training.

**Data availability:** The data is available at https://archive.ics.uci.edu/ml/datasets/Diabetes+130-US+hospitals+for+years+1999-2008#

**Code Availability**: The code will be available after review upon request.

**Declarations Ethics Approval**: No human participants, their data, or biological material was used in this perspective's pieces. Therefore, ethics approval is not applicable in this case.

**Consent to Participate**: Does not apply as no human subjects were involved.

**Consent for Publication** Does not apply as no human subjects were involved.

**Conflicts of Interest** The authors declare no competing interests.